\documentclass[10pt,twocolumn,letterpaper]{article}

\usepackage{iccv}
\usepackage{times}
\usepackage{epsfig}
\usepackage{graphicx}
\usepackage{amsmath}
\usepackage{amssymb}

\usepackage{ulem}
\usepackage{xcolor}
\usepackage{enumitem}
\usepackage{float}
\usepackage{pdfpages}
\usepackage{multirow}
\usepackage{booktabs}
\usepackage{dblfloatfix}
\usepackage{tablefootnote}
\setlist{noitemsep} 
\graphicspath{ {./images/} }
\usepackage[breaklinks=true,bookmarks=false]{hyperref}

\iccvfinalcopy % *** Uncomment this line for the final submission

\def\iccvPaperID{7048} % *** Enter the ICCV Paper ID here

\ificcvfinal\fi

\begin{document}

%%%%%%%%% TITLE
\title{MolGrapher: Graph-based Visual Recognition of Chemical Structures}

\author{
  Lucas Morin\textsuperscript{1, 2} \quad Martin Danelljan\textsuperscript{2} \quad Maria Isabel Agea\textsuperscript{1} \quad Ahmed Nassar\textsuperscript{1}\\
  Valery Weber\textsuperscript{1} \quad Ingmar Meijer\textsuperscript{1} \quad Peter Staar\textsuperscript{1} \quad Fisher Yu\textsuperscript{2}\\
  \textsuperscript{1}IBM Research \quad \textsuperscript{2}ETH Zurich \\
  {\tt\small \{lum, ahn, vwe, inm, taa\}@zurich.ibm.com} \quad {\tt\small martin.danelljan@vision.ee.ethz.ch} \quad {\tt\small i@yf.io} 
}

\newcommand{\taa}[1]{\textcolor{red}{TAA [#1]}}
\newcommand{\inm}[1]{\textcolor{red}{INM [#1]}}
\newcommand{\vwe}[1]{\textcolor{red}{VWE [#1]}}
\newcommand{\isa}[1]{\textcolor{orange}{ISA [#1]}}
\newcommand{\lum}[1]{\textcolor{red}{LUM [#1]}}
\newcommand{\ahn}[1]{\textcolor{teal}{AHN [#1]}}

\ificcvfinal\thispagestyle{empty}\fi

\maketitle

%-------------------------------------------------------------------------
\begin{abstract}

The automatic analysis of chemical literature has immense potential to accelerate the discovery of new materials and drugs. Much of the critical information in patent documents and scientific articles is contained in figures, depicting the molecule structures. However, automatically parsing the exact chemical structure is a formidable challenge, due to the amount of detailed information, the diversity of drawing styles, and the need for training data. In this work, we introduce MolGrapher to recognize chemical structures visually. First, a deep keypoint detector detects the atoms. Second, we treat all candidate atoms and bonds as nodes and put them in a graph. This construct allows a natural graph representation of the molecule. Last, we classify atom and bond nodes in the graph with a Graph Neural Network. To address the lack of real training data, we propose a synthetic data generation pipeline producing diverse and realistic results. In addition, we introduce a large-scale benchmark of annotated real molecule images, USPTO-30K, to spur research on this critical topic. Extensive experiments on five datasets show that our approach significantly outperforms classical and learning-based methods in most settings. Code, models, and datasets are available \footnote{https://github.com/DS4SD/MolGrapher}.

\end{abstract}

%-------------------------------------------------------------------------
\section{Introduction}

\begin{figure}
    \centering
    \includegraphics*[trim={0 0.5cm 0 0cm}, clip, width=0.5\textwidth]{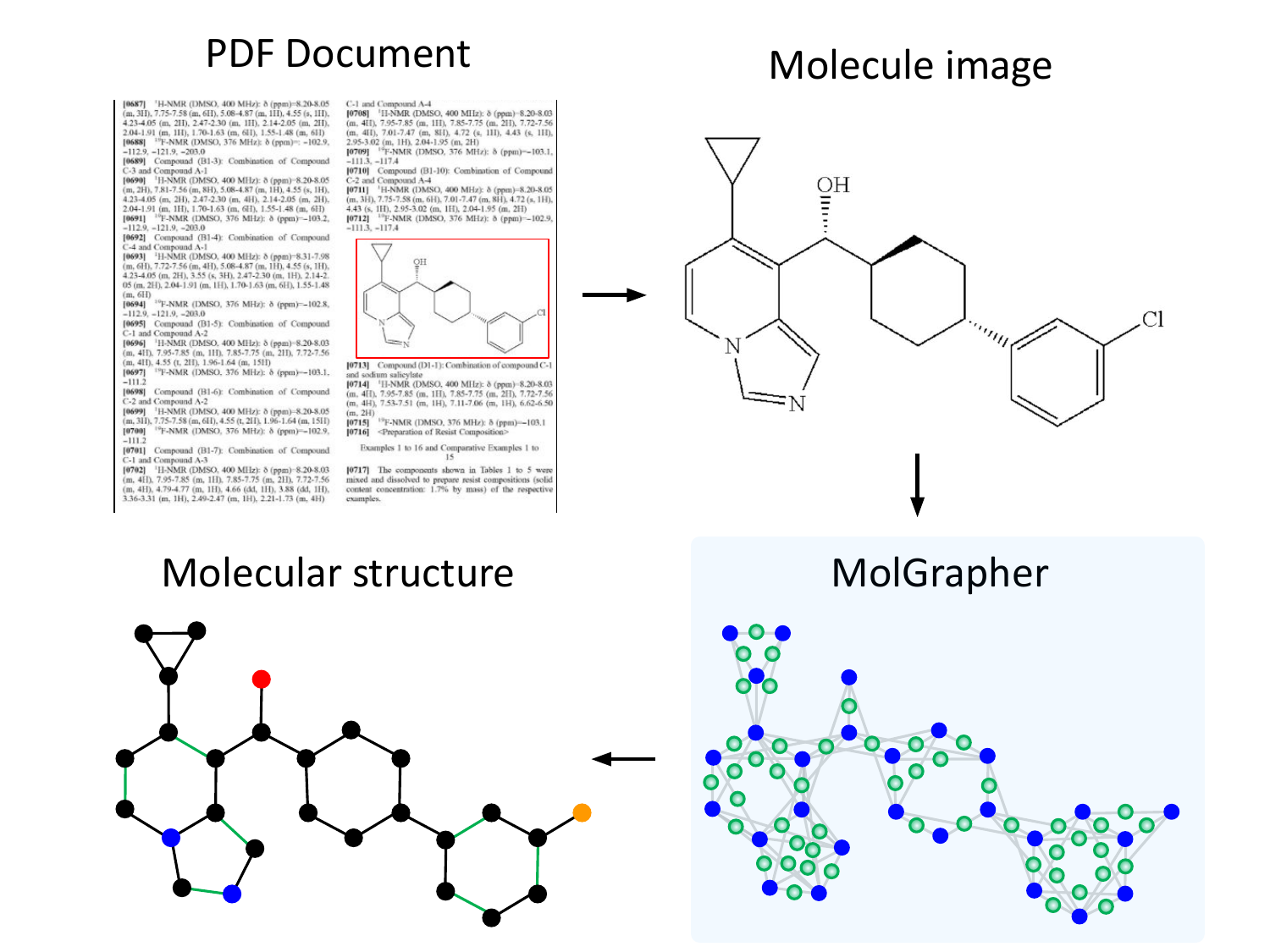}
    \caption{\textbf{MolGrapher} extracts the chemical structure, including all atoms and bonds, from a molecule image in a document. Our approach constructs a supergraph of the molecule (bottom right), containing all detected atom and bond candidates. These nodes are then classified by a Graph Neural Network in order to retrieve the chemical structure.}%
    \label{fig:introduction}\vspace{-2mm}%
\end{figure}

The creation of an open-source database offering a unified view of our current knowledge of all studied molecules, would greatly accelerate research and development in numerous fields, ranging from the pharmaceutical industry to semiconductor manufacturing.
Currently, information about a molecule's properties is distributed across various databases, research articles, and chemical patents, each measuring different subsets of properties.
In such documents, molecules are most often described using images, by drawing their molecular structures. Automatic parsing of molecules from document images, known as Optical Chemical Structure Recognition (OCSR), is thus a critical task in the quest towards establishing a large-scale digital molecule database.

OCSR, however, poses multiple key challenges, which limit the accuracy of current solutions. First, a molecule can be drawn with a variety of styles and conventions. Even the projection of the molecular structure from 3-D onto a 2-D drawing is not unique, leading to the development of various projection algorithms with diverse results. 
Secondly, OCSR requires the extraction of detailed information from the image. Molecules often contain even hundreds of atoms and bonds, which all need to be correctly classified and linked. 
Thirdly, molecules have a virtually endless diversity stemming from the combinatorical explosion of possibilities, presenting particular challenges to data-driven deep learning methods.
Lastly, available real training datasets are extremely limited, further complicating the application of deep learning.

Initial OCSR approaches followed the \emph{graph reconstruction} paradigm. These methods start by extracting the fundamental components of molecules, such as atoms, bonds, and charges, often using hand-crafted image processing algorithms \cite{OSRA, Imago, MolVec}. Then, rules are applied to connect these elements to reconstruct a graph representation of the molecule. 
Some recent works \cite{ChemGrapher, MolMiner, ABC-Net} introduce deep learning elements into the atom and bond detection steps. However, these approaches still rely on hand-crafted rules to link the recognized constituents, and thus not fully benefit from the advantages brought by deep learning. 

Most recently, approaches instead follow the \emph{image captioning} \cite{Img2Mol, DECIMER-AI, SwinOCSR} paradigm by applying a deep network to directly output a character string identifying the molecule, \eg in the form of SMILES~\cite{SMILES}. However, these methods do not explicitly exploit the rich priors and invariances provided by the graph structure. The SMILES representation does not encode the geometrical and neighborhood structure of the molecule graph in a natural manner. While a recent work \cite{Image2Graph} predicts a graph in an autoregressive manner, adding one atom at a time prevents from fully exploiting the graph structure. Partially due to the lack of such a strong inductive bias, image captioning based methods still lag behind from the performance of state-of-the-art rule-based methods \cite{OSRA, Imago, MolVec} on standard benchmarks \cite{OSRA, ChemInfty, MolRec, CLEF}. Moreover, the performance of captioning based methods severely degrade when increasing the size and complexity of the molecule, as they struggle to recover the detailed information in the image.

We introduce MolGrapher for Optical Chemical Structure Recognition, illustrated in Fig.~\ref{fig:introduction}. Our approach explicitly utilizes the graph structure as a strong inductive bias, while also allowing for the final molecule structure itself to be predicted by a deep learning model. MolGrapher operates in three steps. First, we locate atoms and abbreviations in the image with a keypoint detector network. Given the estimated locations of the atoms, we form a \emph{supergraph} of the molecular structure, where two types of nodes represent candidate atoms and bonds respectively (see Fig.~\ref{fig:introduction}). Unlike previous approaches, we do not commit to a final molecule structure at this point, instead extracting a larger set of candidate bonds, forming a superset of the final molecular graph. Candidates that do not correspond to any bond are removed in the final prediction step by including a `no bond' class.

In the final step, we input the constructed supergraph into a Graph Neural Network (GNN). We first extract deep node embeddings through a backbone network. The GNN then operates on the neighborhood structure provided by the supergraph in order to integrate visual information with learned chemistry priors. We then read-out the results with MLPs, classifying each node into a set of atom and bond classes. We also identify abbreviations through a separate class, which are parsed by an OCR component. Since training data is scarce, we develop, and release, a synthetic training data generation pipeline, capable of generating diverse and challenging examples to ensure robust generalization to a wide variety of drawing styles. 

To further aid the research in the community, we introduce USPTO-30K, a large-scale benchmark dataset of annotated real molecule images. It contains separate subsets, in order to independently study the recognition of simple molecules, abbreviated molecules and extremely large molecules. 
We perform comprehensive experiments on five benchmarks: our USPTO-30K, USPTO \cite{OSRA}, Maybridge UoB \cite{MolRec}, CLEF-2012 \cite{CLEF} and JPO \cite{ChemInfty}. We outperform all deep learning based methods that only utilizes synthetic training. Our approach even surpass previous methods that rely on finetuning on real data on most benchmarks, and outperform the longstanding state-of-the-art by rule-based methods \cite{OSRA, Imago, MolVec} in most settings. 

\begin{figure*}[t]
\centering
    \includegraphics*[trim={0 3.8cm 0 3.5cm}, clip, width=\textwidth]{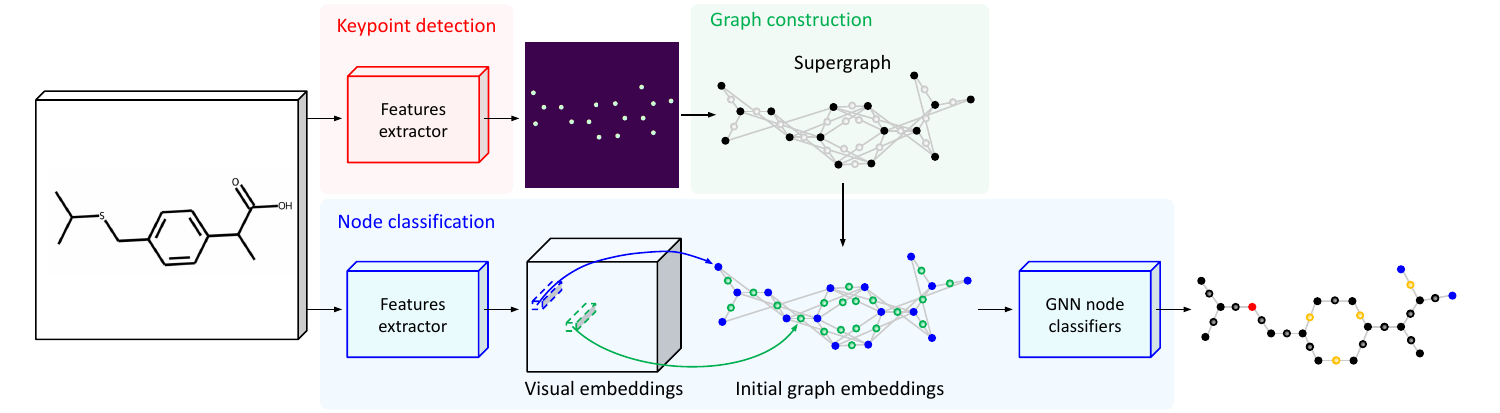}
    \caption{\textbf{Molecular graph recognition architecture.} We illustrate the architecture of MolGrapher, a graph-based network for Optical Chemical Structure Recognition. The keypoint detector (red) locates atoms nodes in the molecule. A supergraph containing atom and bond candidates is constructed (green). Atoms and bonds are classified using a Graph Neural Network (blue).}%
    \label{fig:overview}\vspace{-2mm}%
\end{figure*}

%-------------------------------------------------------------------------
\section{Related work}

OCSR approaches can be divided in two main categories.

\noindent\textbf{Image captioning based OCSR.}
Most of the recent end-to-end deep learning approaches are based on image captioning. These models use an encoder to extract visual features from the image and a decoder to translate them to a SMILES \cite{SMILES} or InChI \cite{InChI} sequence. 
More precisely, DECIMER \cite{DECIMER}, MICER \cite {MICER}, MSE-DUDL \cite{MSE-DUDL} and Img2Mol \cite{Img2Mol} are using a convolutional encoder and a recurrent decoder which is respectively, a GRU decoder, a LSTM decoder, a GridLSTM decoder, and a RNN decoder.
Later works proposed replacing the recurrent decoder with a transformer encoder-decoder, including DECIMER 1.0 \cite{DECIMER-1}, SwinOCSR \cite{SwinOCSR}, IMG2SMI \cite{IMG2SMI} and Image2SMILES \cite{Image2SMILES}.
Other image captioning methods are solely based on transformers, using a vision transformer in \cite{Sundaramoorthy}, a Swin transformer in \cite{Graph-generation} or a Deep TNT transformer encoder with a transformer decoder in ICMDT \cite{ICMDT}. 

One drawback of image captioning methods is the need of large training sets. The SMILES representation has numerous downsides for a learning task \cite{SELFIES}, one of them which is ambiguity due to the association of very different string identifiers to similar molecules. By predicting SMILES, instead of a more faithful representation of the image, the model needs to learn jointly to recognize the image and understand the SMILES language. Multiple string identifiers, possibly suited for a learning task have been proposed, including SELFIES, DeepSMILES \cite{String-idenfiers} or abstract mathematical identifiers \cite{Img2Mol}. Nevertheless, representing a molecule image as a string adds extra complexity for a learning task. Additionally, these models to not properly handle uncertainty by predicting incorrect, but still valid, molecules for challenging inputs such as large molecules. Yet, to parse scientific literature at large scale, avoiding these false positives is critical. This behaviour is explained by the fact that the model is trained to always output valid SMILES, while using global image features. Contrary to these approaches, our model extracts precise visual features for each elementary component of a molecule and operate on a graph, allowing substantially better recognition performance for large molecules, possibly larger than the ones used for training.

\noindent\textbf{Graph reconstruction based OCSR.}
Traditional approaches for Optical Chemical Structure Recognition rely on hand-crafted image processing algorithms to detect elementary components of the molecules and connect them to reconstruct the molecular graph \cite{CSR, OSRA, CLiDE, ChemInk, MolRec, Imago, MolVec}.
More recent works detect elementary components using machine learning models. For instance, ChemGrapher \cite{ChemGrapher} or ABC-Net \cite{ABC-Net} use convolutional segmentation networks while MolMiner \cite {MolMiner} employs a YOLO object detector.
As studied in \cite{CEDe}, detection based methods can be trained with fewer training samples. By doing multiple low-level detections, it is possible to supervise precisely the training. It also offers a natural way to inject prior knowledge, or impose chemistry rules to the model.
Our method benefits from these advantages but differentiates itself by learning both the detection and the association of the fundamental components of molecules. 

Image2Graph \cite{Image2Graph} uses a transformer to build the graph in an auto-regressive way, by adding one atom at the time. \cite{Graph-generation} predicts a SMILES sequence, but enriched with atoms positions, and connectivity information, drifting away from the string identifier and getting closer to a graphical representation.
Our new graph reconstruction approach do not build a graph auto-repressively. We first detects all possible components and associations, and then classify them. It allows to use localized visual features, as well as a complete neighborhood contextual information for each atom and bond prediction.

\section{MolGrapher}

We introduce MolGrapher, a graph-based network for Optical Chemical Structure Recognition.
Our model architecture is illustrated in \autoref{fig:overview}.
Our pipeline consists of three steps. Firstly, the keypoint detector locates atoms nodes in the molecule. Secondly, we construct a graph containing atom and bond candidates. Finally, atoms and bonds are classified using a Graph Neural Network.

\subsection{Keypoint Detection}

In order to construct the molecular graph, we first need to localize keypoints in the image.
Keypoints refer to the position of atoms and `superatoms', \ie abbreviated groups of atoms.
In the later stages, this enables the extraction of relevant visual features for atoms and bonds, as well as the construction of our molecule supergraph.

Our keypoint detector predicts a heatmap, where each peak corresponds to an atom location. 
The output of a feature extractor is passed through convolutional layers to predict the final single-channel heatmap of the atom locations.
To extract the peak locations, we first threshold the heatmap by removing any value in the bottom 10\textsuperscript{th} percentile. We then collect regional maxima, using a window size of $5\times 5$, as the final atom locations. 

Atoms are labeled with Gaussian functions, allowing smoother supervision as well as capturing a degree of uncertainty regarding the predicted positions. Since the ground-truth heatmaps predominantly contains background values, we use the weight-adaptive heatmap regression loss \cite{WAHR} to overcome the class imbalance.
It reduces the impact of well-classified samples, which mostly constitute background, allowing the training to focus on harder samples.

\subsection{Supergraph Construction}

In this section, we describe the construction of the supergraph, which contains all atom and bond candidates to be classified. The supergraph is composed of two types of nodes: atoms and bonds. Further, each edge in the graph only connects an atom node to a bond node. Representing both of these entities as nodes allows us to process them uniformly through standard message passing layers. 

The input to our supergraph construction is a set of approximate atom locations. Since the connectivity between the detected atoms is unknown, our graph has to represent a superset of the final molecule prediction, thus termed `supergraph'. That is, it contains all possible bonds. By including a prediction class corresponding to `no bond', bond candidates which are not actually in the molecule can be removed in the later stage. Including an analogous class for `no atom', our approach can even correct false positives predicted by the keypoint detection network. 

\begin{figure}[t]
    \includegraphics[trim={0 2cm 0 2cm}, clip, width=0.5\textwidth]{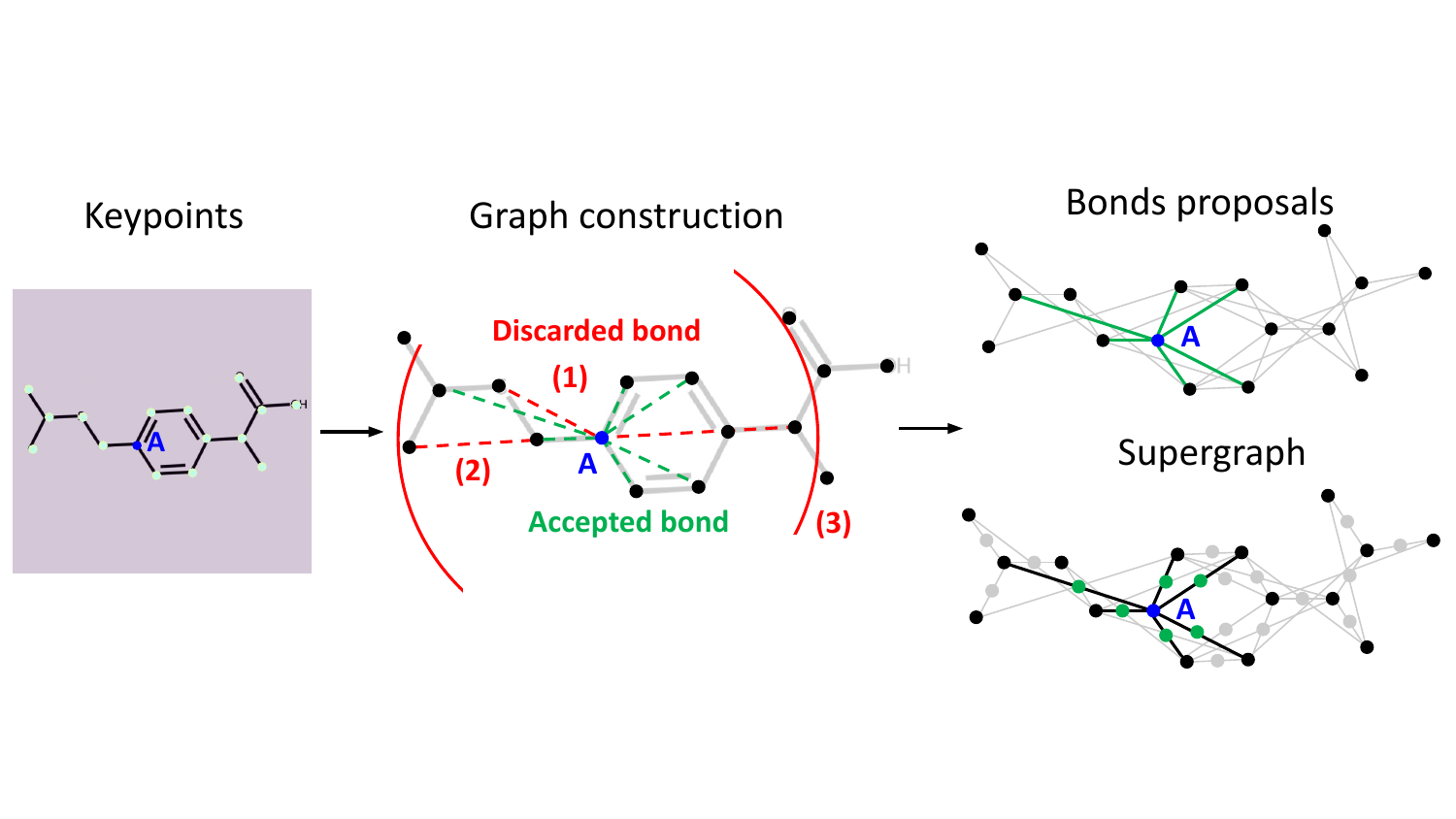}
    \caption{\textbf{Supergraph construction.} The figure presents the construction of bonds proposals for an atom denoted $A$. Considered bonds are depicted with dashed lines. Green bonds are accepted in the supergraph, while red bonds are discarded because: (1) there are no filled pixels around their centerpoints or (2) they are obstructed by other keypoints.}
    \label{fig:construction}\vspace{-2mm}
\end{figure}

Our supergraph construction is illustrated in \autoref{fig:construction}.
We first consider each keypoint individually, adding connections to all other keypoints within a radius of three times the estimated bond length \cite{OSRA}. Subsequently, we remove connections which have no filled pixels around their centerpoints or if there is a third keypoint located between its two extremities. 
Finally, we keep at most six bond candidates for each atom, by eliminating the longest connections. Chemistry prior knowledge ensures that it is extremely unlikely for an atom to have more than five neighboring bonds. Our settings are highly conservative to avoid deleting any actual bond. On the other hand, pruning superfluous bonds avoids any unnecessary false positives in the later stage and a more effective connectivity for the GNN.

\subsection{Node Classification}

In order to classify atom and bond nodes, we integrate a Graph Neural Network (GNN) that operates on the constructed supergraph.
The embeddings for all nodes are initialized using visual features as well as their types. This approach leverages the local context of atoms and bonds to predict their class. Intuitively, it enables the learning of chemistry rules, \ie atoms have a fixed valence, as well as chemistry prior knowledge, \ie some typical substructures are more common than others. 

Let $e_{i}^{1} \in R^{D}$ denotes the initial embedding of the node $i$ of type $t_i \in \left\{\text{atom}, \text{bond}\right\}$ located at position $(x_{i}, y_{i})$. It is computed by combining a visual feature vector $v_i$ and a learnable type encoding $w_{t_i}$,
\begin{equation}
e_{i}^{0} = v_{i} + w_{t_i}
\end{equation}
The visual features $v_{i}$ are extracted from a deep feature map $F$, extracted by a backbone network. For an atom node $i$ we evaluate the feature map in the corresponding location $(x_{i}, y_{i})$ with bilinear interpolation,
\begin{equation}
    v_{i} = F(x_{i}, y_{i}) \,.
\end{equation}
For a bond node $i$, we aggregate features at locations along the bond as
\begin{equation}
    v_{i} = \frac{1}{J}\sum_{j=1}^{J} F(x_i^{j}, y_i^{j}) \,,
\end{equation}
where $(x_i^{j}, y_i^{j}) $ are the locations of $J$ uniformly spaced points along the bond itself.

Given initial node embeddings, information is propagated through the graph to learn local dependencies between atoms and bonds. Indeed, directly neighboring atoms and bonds respect strict chemistry rules, and larger neighborhoods form patterns called functional groups, which are typically abundant in nature. 
We employ GNN layers, denoted $\left\{g^{k}\right\}_{k \in [1, N]}$, to iteratively update the node embeddings,
\begin{equation}
e^{k+1} = g^{k}(e^{k}) \,.
\end{equation}
The final predictions $\left\{p_{i}\right\}$ are then obtained using two Multi-Layer Perceptrons, one for each node type $t$, 
\begin{equation}
p_{i} = MLP_{t}(e_{i}^{N}) \,.
\end{equation}
The vector $p_{i} \in \mathbb{R}^{C_t}$ contains the logits for the final atom or bond classes. 

The most represented bond types include `no bond', `single', `double' and `triple'. Atom classes include the common (\eg O) and exotic atoms as well as charged atoms (\eg O$^-$). Since molecule images often include whole groups in the nodes, \eg COOH, we further include a class `superatom' to recognize these cases. Such nodes are then parsed as described in Section~\ref{section:Superatom-groups-recognition}.

Our node classification model is trained using the cross-entropy loss for both the atom and bond prediction heads. During training, the supergraph structure used as input is obtained by applying the graph construction algorithm on augmented ground truth keypoints.
To simulate possible errors from the keypoint detector, we add noise to the atom positions and randomly include false-positive keypoints.
Decoupling the keypoint detection and the node classification allow to train faster and to use optimized training sets for each task.

\subsection{Superatom Groups Recognition}
\label{section:Superatom-groups-recognition}

To recognize abbreviated substructures, named superatoms, we use an external Optical Character Recognition (OCR) system.
Relying on an external OCR system provides strong robustness, since the model is trained on a diversity of text representations that cannot necessarily be generated in a molecule image.
At the location of an abbreviated group, the node classifier predicts a `superatom' class. Then, the OCR engine PP-OCR \cite{PaddleOCR} is used to recognize the text written at this location, which is finally replaced by its corresponding sub-molecule, stored in a precomputed mapping.  

\section{Datasets}

\subsection{Synthetic Training Data Generation}
\label{section:Synthetic training set}

Training datasets of annotated molecule images extracted from scientific publications are extremely limited. Therefore, we develop a synthetic training pipeline to generate a broad diversity of molecular representations. It further allows us to extract atom-level annotations. 

Our dataset is created using molecule SMILES retrieved from the database PubChem \cite{PubChem}. To increase the probability that the dataset covers the largest variety of molecular structures, we sample SMILES using a strategy based on functional groups. A functional group is molecule substructure typically abundant in nature. Thus, we retrieve molecules from PubChem containing each functional group in a defined collection of 1540 functional groups. To further increase diversity, these molecules are queried with different ranges of number of atoms. Finally the dataset is filtered by removing salts, molecules containing complex polycyles, isotopes or radical electrons. 
Training images are then generated from SMILES using the molecule drawing library RDKit \cite{RDKit}.

To capture the large diversity of drawing styles and conventions from different scientific documents, the synthetic set is highly augmented, at multiple levels. 
Firstly, molecules are randomly transformed, notably by setting the selection of a molecular conformation, adding artificial superatom groups with single or multiple attachment points or displaying solid, dashed or wavy bonds. Secondly, the rendering parameters used in RDKit are randomly set, including the font, the bond width, the display of aromatic cycles using circles. This leads to substantial variations in the drawn molecule, to aid generalization to real datasets.

The generated molecule images are saved together with their corresponding MolFiles. A MolFile \cite{MolFile} stores information about the atoms and their positions in a molecule, as well as bonds and their connectivity. It provides the necessary information for supervising the training.
This generation pipeline covers a large diversity of molecules including stereo-chemistry and molecules with superatom groups. Finally, the images undergo several image augmentations on the fly such as gaussian blurring, the adding of pepper patches, random lines, and random captions.

The exhaustive lists of molecule, rendering and image augmentations, as well as details regarding the generation of atom-level annotations are available in the supplementary materials.

%-------------------------------------------------------------------------
\begin{table*}[!t]
\centering%
\caption{\textbf{Comparison of our method with existing OCSR models.} We report the accuracy, \ie the percentage of perfectly recognized molecule images, on datasets coming from real scientific documents. $\ast$: re-implemented results. \textdagger: results from original publications. \textdaggerdbl: results from \cite{CEDe}.}
\label{tab:results}
\resizebox{0.90\linewidth}{!}{%
\begin{tabular}{lccccccc}
\toprule
\textbf{Method} & \begin{tabular}[c]{@{}c@{}}USPTO\\ (5719)\end{tabular} & \begin{tabular}[c]{@{}c@{}}Maybridge UoB\\ (5740)\end{tabular} & \begin{tabular}[c]{@{}c@{}}CLEF-2012\\  (992)\end{tabular} & \begin{tabular}[c]{@{}c@{}}JPO \\ (450)\end{tabular} & \begin{tabular}[c]{@{}c@{}}USPTO-10K\\ (10 000)\end{tabular} & \begin{tabular}[c]{@{}c@{}}USPTO-10K-Abb\\ (10 000)\end{tabular} & \begin{tabular}[c]{@{}c@{}}USPTO-10K-L\\ (10 000)\end{tabular} \\ \hline
\textit{Rule-based methods} \\
OSRA 2.1 \cite{OSRA} \textsuperscript{$\ast$} & 89.3 & 86.3 & \textbf{93.4} & 56.3 & 89.7 & 63.9 & 43.1 \\
MolVec 0.9.7 \cite{MolVec} \textsuperscript{$\ast$} & 89.1 & 88.3 & 81.2 & 66.8 & 92.4 & 70.3 & \textbf{64.0} \\
Imago 2.0 \cite{Imago} \textsuperscript{$\ast$} & 89.4 & 63.9 & 68.2 & 41.0 & 89.9 & 63.0 & 47.3 \\ \hline
\textit{Only synthetic training} \\
DECIMER 2.0 \cite{DECIMER-AI} \textsuperscript{\textdagger} & 61.0 & 88.0 & 72.0 & 64.0 & - & - & - \\
Image2Graph \cite{Image2Graph} \textsuperscript{\textdagger} & 44.9 & 72.0 & 37.8 & 24.0 & - & - & - \\
Graph Generation \cite{Graph-generation} \textsuperscript{\textdagger}& 67.0 & 83.1 & 74.6 & \textit{-} & - & - & - \\
CEDe \cite{CEDe} \textsuperscript{\textdaggerdbl} & 79.0 & 74.1 & 68.0 & 49.4 & - & - & - \\
ChemGrapher \cite{ChemGrapher} \textsuperscript{\textdaggerdbl} & 80.9 & 83.2 & 75.5 & 53.3 & - & - & - \\
Img2Mol \cite{Img2Mol} \textsuperscript{$\ast$} & 25.2 & 68.0 & 17.9 & 16.1 & 35.4 & 13.8 & 0.0 \\
\textbf{MolGrapher (Ours)} & \textbf{91.5} & \textbf{94.9} & 90.5 & \textbf{67.5} & \textbf{93.3} & \textbf{82.8} & 31.4 \\
\bottomrule
\end{tabular}}\vspace{-1mm}
\end{table*}
%-------------------------------------------------------------------------

\subsection{The USPTO-30K Dataset}
Existing benchmarks have some limitations. Being created using only a few documents, they contain batches of very similar molecules. For example in a patent, a molecule could typically be displayed together with all the substituent of one particular substructure, resulting in large batches of almost identical molecules. Additionally, the existing sets contain molecules of different kinds, including superatom groups and various markush \cite{Markush} features, which should be evaluated independently. In practice, it is important to delimit on which types of molecules models can be applied.

We introduce USPTO-30K, a large-scale benchmark dataset of annotated molecule images, which overcomes these limitations. It is created using the pairs of images and MolFiles \cite{MolFile} by the United States Patent and Trademark Office (USPTO) \cite{USPTO}. Each molecule was independently selected among all the available images from 2001 to 2020. The set consists of three subsets to decouple the study of clean molecules, molecules with abbreviations and large molecules. USPTO-10K contains 10,000 clean molecules, \ie without any abbreviated groups. USPTO-10K-abb contains 10,000 molecules with superatom groups. USPTO-10K-L contains 10,000 clean molecules with more than 70 atoms. We provide visualizations and analysis in the supplementary materials.

\section{Experiments}

In this section, we perform comprehensive experiments on multiple OCSR benchmarks. 
Moreover, we propose an analysis of the different components of our method. 

\subsection{Implementation details}

The implementation is done with PyTorch 1.12 with CUDA 11.3. 
For the keypoint detector features extractor, we use a ResNet-18 backbone with $8\times$ dilatation factor \cite{dilated-cnn} to preserve a high spatial resolution.
For the node classifier features extractor, we resort to a ResNet-50 with a $2\times$ dilation factor.
The model is trained on 300,000 synthetic images, presented in \autoref{section:Synthetic training set}. We train for 20 epochs on 3 NVIDIA A100 GPUs using ADAM with a learning rate of 0.0001 that we decay after 5000 iterations by a factor of 0.8. The losses for atoms and bonds classifiers are weighted by factors 1 and 3, respectively.
During inference, we pre-process images by removing captions using PP-OCR \cite{PaddleOCR}. Additionally, in case an initial prediction is an invalid molecule, we use PP-OCR to merge keypoints located in a same detected text cell. See the supplementary material for more details.

\subsection{Evaluation datasets and metrics}
To compare our method with state-of-the-art, the model is evaluated on the standard benchmarks USPTO \cite{OSRA}, Maybridge UoB \cite{MolRec}, CLEF-2012 \cite{CLEF} and JPO \cite{ChemInfty}. 
USPTO and CLEF are collections of 5,719 and 992 molecule images. Besides, JPO contains 450 images published by the Japanese Patent Office (JPO). JPO is particularly challenging because of its non-standard drawing conventions and poor images qualities. Lastly, Maybridge UoB is a dataset of 5,740 scanned molecule images taken from a catalogue of drug compounds by Maybridge. To compare methods, we compute accuracy, defined as the percentage of perfectly recognized molecules. In practice, this is done by verifying that the predicted and ground-truth molecules have identical InChI \cite{InChI} keys. For evaluation, stereo-chemistry is removed and markush structures are not considered.

%-------------------------------------------------------------------------
\begin{figure*}[!b]
    \includegraphics[trim={0 4.7cm 0 5cm}, clip, width=\textwidth]{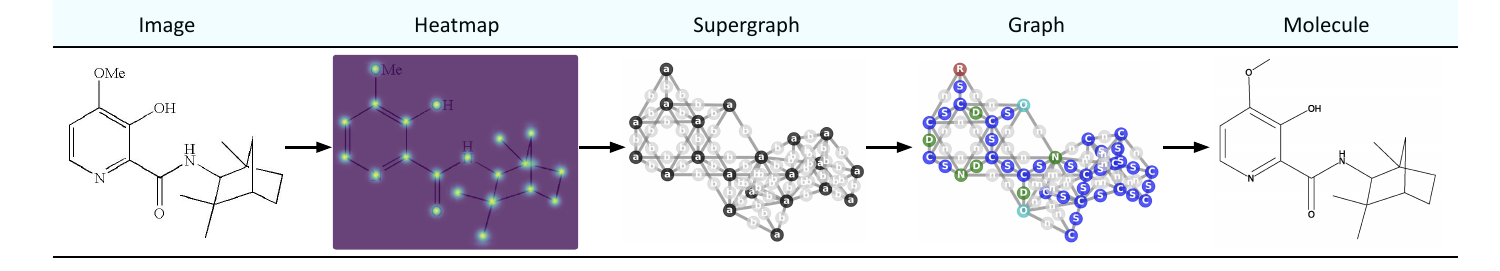}
    \caption{\textbf{Prediction steps.} Keypoints are detected and then used to build a supergraph. After classifying the nodes of the graph and recognizing abbreviated groups, the output molecule is created. In this example, the polycyclic molecule contains overlapping bonds, a challenging feature for OCSR models, and is still correctly recognized.}
    \label{fig:inference}
\end{figure*}
%-------------------------------------------------------------------------

\subsection{State-of-the-art Comparison}

\autoref{tab:results} compares the OCSR methods on different benchmarks. 
Our method achieves superior results compared to rule-based models on most datasets, including USPTO, Maybridge UoB and JPO. 
Solely using synthetic training, our method reduces the error rate by more than half compared to other deep learning methods on USPTO, Maybridge UoB, and CLEF-2012. 
The design of our model confers a strong generalization ability. In practice, this is very important as our model will be applied to a wide variety of documents, which are not necessarily evaluated by the current benchmark datasets. MolGrapher also maintains good performance for extremely large molecules in USPTO-10K-L, which only contains examples with more than 70 atoms. On the contrary, recognizing large molecules is a significant challenge for image captioning based methods. For instance, Clevert \etal point out that the performance of Img2Mol \cite{Img2Mol} drop sharply for molecules larger than 40 atoms. This is confirmed in our evaluation, where it fails to correctly recognize virtually any molecule in USPTO-10K-L. \autoref{tab:finetuning} also provides comparison with models finetuned using real data. Our model surpasses these methods on Maybridge UoB, CLEF-2012 and JPO and is second best on USPTO, while not relying on any real data for training. 

Our model outperforms existing methods on our USPTO-10K and USPTO-10K-Abb. The performance is particularly improved for molecules containing abbreviated groups. The performance gap between USPTO and USPTO-10K-Abb for rule-based approaches suggests that these methods are specifically parameterized for abbreviations in the existing benchmark datasets. Our model is the only deep learning method to provide a competitive performance for extremely large molecules. This is allowed by our architecture, which extracts localized visual features. 

%-------------------------------------------------------------------------
\begin{table}[t]
\centering% 
\caption{\textbf{Comparison of our method with deep OCSR models finetuned on real data.} We report the accuracy, \ie the percentage of perfectly recognized molecule images, on datasets coming from real scientific documents. \textdagger: results from original publications. \textdaggerdbl: results (in {\color[HTML]{9B9B9B} grey}) from unavailable sub-splits of the original benchmarks, only reported for reference.}
\label{tab:finetuning}
\resizebox{\linewidth}{!}{%
\begin{tabular}{lcccc}
\hline
\textbf{Method} & \begin{tabular}[c]{@{}c@{}}USPTO\\ (5719)\end{tabular} & \begin{tabular}[c]{@{}c@{}}Maybridge UoB\\ (5740)\end{tabular} & \begin{tabular}[c]{@{}c@{}}CLEF-2012\\  (992)\end{tabular} & \begin{tabular}[c]{@{}c@{}}JPO \\ (450)\end{tabular} \\ \hline
\textit{Real data finetuning} & \multicolumn{1}{l}{} & \multicolumn{1}{l}{} & \multicolumn{1}{l}{} & \multicolumn{1}{l}{} \\
Image2Graph \cite{Image2Graph} \textsuperscript{\textdagger} & 55.1 & 83.0 & 51.7 & \underline{50.0} \\
Graph Generation \cite{Graph-generation} \textsuperscript{\textdagger} & \textbf{92.9} & \underline{86.6} & \underline{87.5} & - \\ 
CEDe \cite{CEDe} \textsuperscript{\textdagger} \textsuperscript{\textdaggerdbl} & {\color[HTML]{9B9B9B} 91.0} & {\color[HTML]{9B9B9B} 91.5} & {\color[HTML]{9B9B9B} 86.6} & {\color[HTML]{9B9B9B} 74.0} \\ \hline
\textit{Only synthetic training} & \multicolumn{1}{l}{} & \multicolumn{1}{l}{} & \multicolumn{1}{l}{} & \multicolumn{1}{l}{} \\
\textbf{MolGrapher (Ours)} & \underline{91.5} & \textbf{94.9} & \textbf{90.5} & \textbf{67.5} \\ \bottomrule
\end{tabular}} \vspace{-3mm}
\end{table}
%-------------------------------------------------------------------------

\subsection{Model Robustness}

Clevert \textit{et al.} \cite{Img2Mol} introduced modified version of the standard benchmarks by applying slight perturbations, such as rotation and shearing, to the input images. \autoref{tab:perturbed} presents a comparison to previous methods on the same perturbed sets. The augmented images can be seen as a simulation of scanned images, which are typically found in chemical patents.
We observe that the performance of rule based approaches, OSRA, MolVec and Imago, decreases sharply. However, our model maintains a good result, outperforming other methods by a significant margin. This robustness is essential for large-scale analysis of chemical literature, as input molecule image quality cannot be controlled.
%-------------------------------------------------------------------------
\begin{table}{}
\centering
\caption{\textbf{Evaluation of model robustness.} We report the accuracy, \ie the percentage of perfectly recognized molecule images, on perturbed datasets coming from real scientific documents. \textdagger: results from \cite{Img2Mol}.}
\label{tab:perturbed}
\resizebox{\linewidth}{!}{
\begin{tabular}{lcccc}
\toprule
\textbf{Method} & \begin{tabular}[c]{@{}c@{}}USPTO\textsuperscript{p}\\ (4852)\end{tabular} & \begin{tabular}[c]{@{}c@{}}Maybridge UoB\textsuperscript{p}\\ (5716)\end{tabular} & \begin{tabular}[c]{@{}c@{}}CLEF-2012\textsuperscript{p}\\  (711)\end{tabular} & \begin{tabular}[c]{@{}c@{}}JPO\textsuperscript{p} \\ (365)\end{tabular} \\ \midrule
OSRA 2.1 \cite{OSRA} \textsuperscript{\textdagger} & 6.4 & 70.9 & 17.0 & 33.0 \\
MolVec 0.9.7 \cite{MolVec} \textsuperscript{\textdagger} & 30.7 & 75.0 & 44.5 & 49.5 \\
Imago 2.0 \cite{Imago} \textsuperscript{\textdagger} & 5.1 & 5.1 & 26.7 & 23.2\\
Img2Mol \cite{Img2Mol} \textsuperscript{\textdagger} & 42.3 & 78.2 & 48.8 & 45.1 \\
\textbf{MolGrapher (Ours)} & \textbf{86.7} & \textbf{94.1} & \textbf{87.8} & \textbf{55.4} \\
\bottomrule  
\end{tabular}} \vspace{-2mm}
\end{table}
%-------------------------------------------------------------------------

\subsection{Qualitative evaluation}

\begin{figure*}[t]
\vspace{-2mm}
\centering%
    \includegraphics[trim={0 0cm 0 0cm}, clip, width=\textwidth]{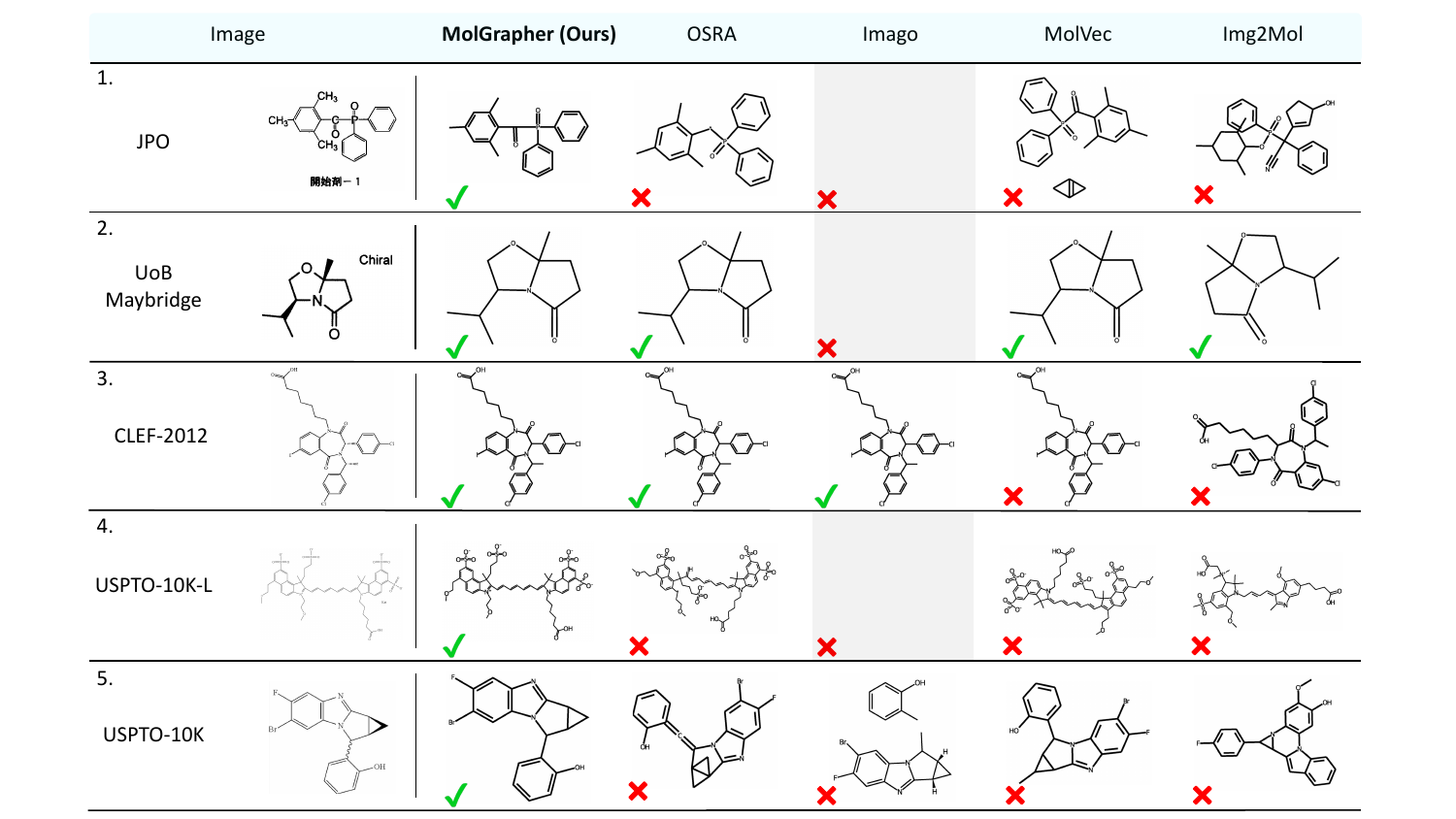}%
    \caption{\textbf{Qualitative comparison.} The figure shows examples of predictions for characteristic images from different benchmarks. Compared to previous rule-based and learning based methods, our approach robustly recognizes the exact molecular structure in challenging cases, such as with distracting captions, stereo-chemistry, and very large molecules.}%
    \label{fig:comparison}\vspace{-2mm}
\end{figure*}

In this section, we conduct a qualitative evaluation of MolGrapher, in comparison to state-of-the-art methods. First, \autoref{fig:inference} illustrates the intermediate outputs of MolGrapher. Note that MolGrapher accurately recognizes molecules in challenging cases with overlapping bonds, without the use of any post-processing rules. \autoref{fig:comparison} showcases examples of predicted molecules for images from various benchmark datasets. MolGrapher demonstrates robustness to images containing captions (\autoref{fig:comparison} row 1). It can correctly recognize extremely large molecules (\autoref{fig:comparison} row 4), and is robust to solid, dashed or wavy bonds (\autoref{fig:comparison} row 5). Unlike image captioning based methods such as as Img2Mol, our predictions preserve the projection and atom placement used in the input image (\autoref{fig:comparison} row 2). This can convey critical information for human interpretation. 

%-------------------------------------------------------------------------
%-------------------------------------------------------------------------

\subsection{Ablation study}
We conduct an ablation study to demonstrate the impact of each component in the proposed method.

\noindent\textbf{Synthetic training set.} We evaluate the impact of the three levels of augmentation of the training set, illustrated in \autoref{tab:augmentations}. The image augmentations have a major impact on the performance of the model. In practice, we noticed that the keypoint detector benefits greatly from using heavy image transformations. Additionally, the molecule augmentations, which allow training molecules to contain abbreviated groups, proved crucial to handle images containing abbreviated groups, such as in USPTO.

%-------------------------------------------------------------------------
\begin{table}[]
\caption{\textbf{Training set analysis.} Impact of training set augmentations on performance. Molecule, rendering and image level augmentation are independently removed.}
\label{tab:augmentations}
\centering
\resizebox{0.9\linewidth}{!}{
\begin{tabular}{lccc}
\toprule
\multicolumn{1}{l}{} & USPTO & JPO & JPO\textsuperscript{p}\\ \hline
All augmentations & \textbf{91.5} & \textbf{67.5} & \textbf{55.4} \\
No molecule augmentation & 52.4 & 52.7 & 43.4 \\
No rendering augmentation & 78.4 & 55.7 & 44.1 \\
No image augmentation & 69.8 & 29.3 & 22.5 \\ \bottomrule
\end{tabular}}\vspace{-3mm}
\end{table}
%-------------------------------------------------------------------------

%-------------------------------------------------------------------------
\begin{table}[b]\vspace{-4mm}
\caption{\textbf{Keypoint detector analysis.} We analyse the standard deviation of the Gaussian functions (top), as well as the training loss function (bottom). $b$ denotes the molecule bond length.}
\label{tab:analysis-keypoints}
\centering
\resizebox{0.9\linewidth}{!}{
\begin{tabular}{llccc}
\toprule
\multicolumn{2}{l}{} & USPTO & JPO & JPO\textsuperscript{p}\\ \hline
Heatmap standard deviation & $b/5$ & 81.3 & 54.7 & 46.8\\
 & $b/10$ & \textbf{91.5} & \textbf{67.5} & \textbf{55.4} \\
 & $b/15$ & 91.4 & 64.7 & 51.6\\
 \midrule
Training loss & L2 & 91.1 & 66.2 & 55.1\\
 & WAHR \cite{WAHR} & \textbf{91.5} & \textbf{67.5} & \textbf{55.4} \\ \bottomrule
  \\
\end{tabular}}\vspace{-4mm}
\end{table}
%-------------------------------------------------------------------------
\noindent\textbf{Keypoint detector.} In \autoref{tab:analysis-keypoints}, we experiment with different standard deviations for the Gaussian keypoint labels, and keypoint regression losses. Using a large standard deviation of $b/5$, where $b$ denotes the bond length, creates overlaps between keypoints and significantly decreases performance. While the model is not sensitive to the choice of training loss, the WAHR loss offers a slight advantage. 

\noindent\textbf{Supergraph construction.} In \autoref{tab:analysis-classifier}, we evaluate the impact of the maximum number of bond candidates and the search radius in the supergraph construction. We observe that our approach is not sensitive to these settings, but only experiences a small drop when increasing the number of bond candidates. This drop is explained by false positives predicted during the node classification stage.

%-------------------------------------------------------------------------
\begin{table}[t]
\caption{\textbf{Supergraph construction analysis.} We analyse the maximum number of bond candidates (top), and the maximum search radius (bottom). $b$ denotes the molecule bond length.}
\label{tab:analysis-supergraph}
\centering
\resizebox{0.9\linewidth}{!}{
\begin{tabular}{llccc}
\toprule
\multicolumn{2}{l}{} & USPTO & JPO & JPO\textsuperscript{p}\\ \hline
Number bond candidates & 6 & \textbf{91.5} & \textbf{67.5} & \textbf{55.4} \\
 & 10 & 91.0 & 65.7 & 54.1\\
 \midrule
Search radius & $3b$ & \textbf{91.5} & \textbf{67.5} & \textbf{55.4}\\
 & $5b$ & 91.3 & 65.0 & 53.1 \\ \bottomrule
\end{tabular}}\vspace{-3mm}
\end{table}
%-------------------------------------------------------------------------
%-------------------------------------------------------------------------
\begin{table}[t]\vspace{1mm}
\caption{\textbf{Node classifier analysis.} We analyse the features representing graph nodes (top), and the usage of GCN layers (bottom). \textdagger: without abbreviated molecules and charges.}
\label{tab:analysis-classifier}
\centering
\resizebox{0.9\linewidth}{!}{
\begin{tabular}{llccc}
\toprule
\multicolumn{2}{l}{} & USPTO & JPO & JPO\textsuperscript{p}
\\ \hline
Node embedding & Visual & 90.0 & 64.7 & 54.9\\
\multicolumn{1}{c}{} & \hspace{3pt} + Type & \textbf{91.5} & \textbf{67.5} & \textbf{55.4} \\
\multicolumn{1}{c}{} & \hspace{3pt} + Position & 91.3 & 62.8 & 54.8 \\ 
 \toprule
\multicolumn{2}{l}{} & USPTO\textsuperscript{\textdagger} & JPO\textsuperscript{\textdagger} & JPO\textsuperscript{p\textdagger} 
\\  \hline
Number of GCN Layers & 0 & 90.1 & 58.0 & 53.6\\
\multicolumn{1}{c}{} & 4 & \textbf{90.4} & \textbf{66.6} & \textbf{61.6} \\
\bottomrule
\end{tabular}}\vspace{-3mm}
\end{table}
%-------------------------------------------------------------------------

\begin{figure}[t]
\centering%
    \includegraphics[trim={0 3cm 0 4cm}, clip, width=0.5\textwidth]{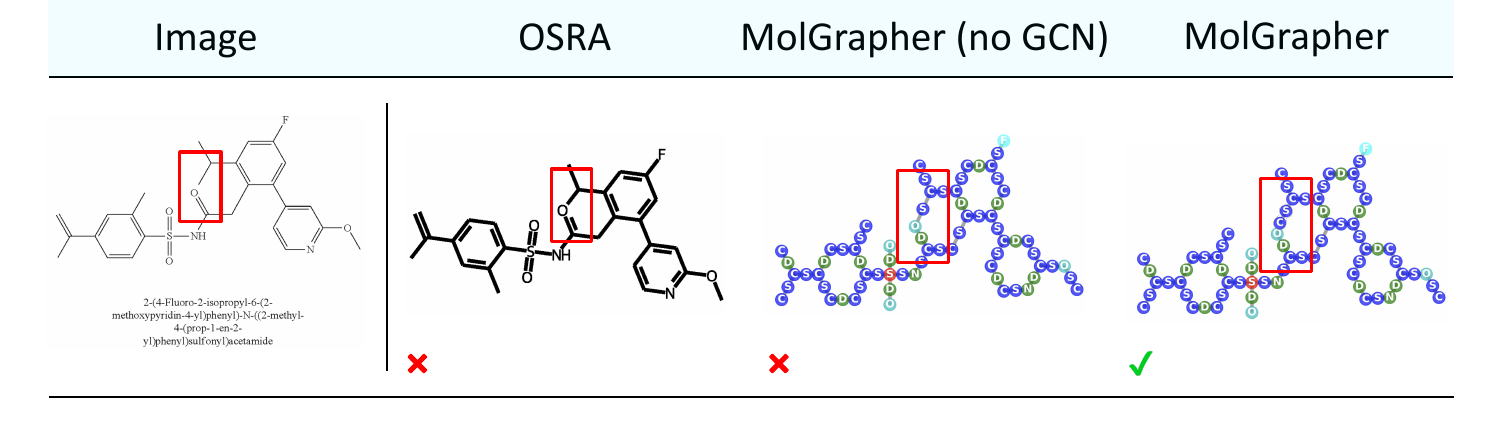}
    \caption{\textbf{Ablation experiment.} The figure shows MolGrapher predictions with and without using GCN layers, illustrating that GCN layers allows the model to learn chemistry rules. Indeed, the model correctly connect the oxygen atom highlighted in red to a double bond only.}%
    \label{fig:gcn}\vspace{-3mm}
\end{figure}

\noindent\textbf{Node classifier.}  As demonstrated in \autoref{tab:analysis-classifier}, initializing nodes embeddings with both visual and type encoding increases performance. We also experimented with positional embedding but found it to reduce the overall performance. 
The model demonstrating, in average, best performances does not include GCN layers. However, as demonstrated in \autoref{tab:analysis-classifier}, for a limited evaluation setup, GCN layers have positive impact on performance, leading to significant gains of 8.6\% and 8.0\% on JPO and JPO\textsuperscript{p} when using 4 instead of 0 GCN layers. Note that even the version using 0 GNN layers, still employs our supergraph for feature aggregation and classification. The propagation of information through the graph is beneficial for challenging cases, such as the one illustrated in \autoref{fig:gcn}. 
In this example, interpreting only the visual information would be misleading, even for humans. Understanding chemistry rules is mandatory to resolve the ambiguities in the drawing, which is correctly done by our model.

%-------------------------------------------------------------------------

\section{Conclusion}

We propose a novel architecture for recognizing 2-D molecule depictions in documents by exploiting the natural graph representation of molecules and graph symmetries. Our model accurately detects atoms and bonds based on their visual features and local context, outperforming existing methods on standard benchmarks and our new USPTO-30K dataset. The model is trained on synthetic images and demonstrates strong generalization capabilities, making it useful at scale without requiring annotations for each journal or patent office. 

\newcommand*{\dictchar}[1]{
    \clearpage
    \twocolumn[
    \centerline{\parbox[c][3.5cm][c]{18cm}{
            \centering
            \fontsize{15}{15}
            \selectfont
            {#1}}}]
}

\begin{center}
\dictchar{\textbf{Supplementary materials for \break MolGrapher: Graph-based Visual Recognition of Chemical Structures\vspace{0.5cm}}}
\end{center}

Here, we provide additional details and visualizations regarding: the synthetic training set in \autoref{sec:training}, the benchmark dataset USPTO-30K in \autoref{sec:uspto-30k}, the MolGrapher implementation in \autoref{sec:implementation}, and the MolGrapher evaluation in \autoref{sec:analysis}. 

\begin{table}[b]
\centering% 
\caption{\textbf{USPTO-30K comparison with other benchmarks.} We compare the number of samples, the number of classes, the molecule sizes and the proportion of molecules with stereo-chemistry.}
\label{tab:comparison}
\resizebox{\linewidth}{!}{%
\begin{tabular}{lcccccccc}
\toprule
\multirow{2}{*}{Dataset} & Number of & \multicolumn{2}{c}{Molecule sizes} & Stereo-chemistry & \multicolumn{3}{c}{Number of classes} \\ \cline{3-4} \cline{6-8} 
 & samples & [Min, Max] & Mean & proportion & Atom & Bond & Superatom \\ \hline
USPTO & 5719 & [10, 96] & 28  & 20.5 & 24 & 6 & 234  \\
Maybridge UoB & 5740 & [4, 34] & 2.6 & 13  & 20 & 6 & 0  \\
CLEF & 992 & [4, 42] & 26 & 33.9 & 15 & 6 & 43  \\
JPO & 450 & [5, 43] & 20 & 1.0 & 9 & 6 & 15  \\ \hline
\textbf{USPTO-30K} & \textbf{30000} & \textbf{[4, 543]} & \textbf{53} & \textbf{39.2} & \textbf{74} & \textbf{6} & \textbf{620} \\
USPTO-10K & 10000 & [4, 198] & 31 & 29.3 & 37 & 6 &  0 \\
USPTO-10K-Abb & 10000 & [4, 162] & 31  & 31.2 & 45 & 6 &  620 \\
USPTO-10K-L & 10000 & [71, 543] & 96 & 57.1 & 52 & 6  & 0 \\ \bottomrule
\end{tabular}}\vspace{-2mm}
\end{table}

\section{Synthetic Training Set details}
\label{sec:training}

\begin{figure*}[b]
\centering
\resizebox{\textwidth}{!}{
    \includegraphics*[trim={0 0cm 0 0cm}, clip, width=\textwidth]{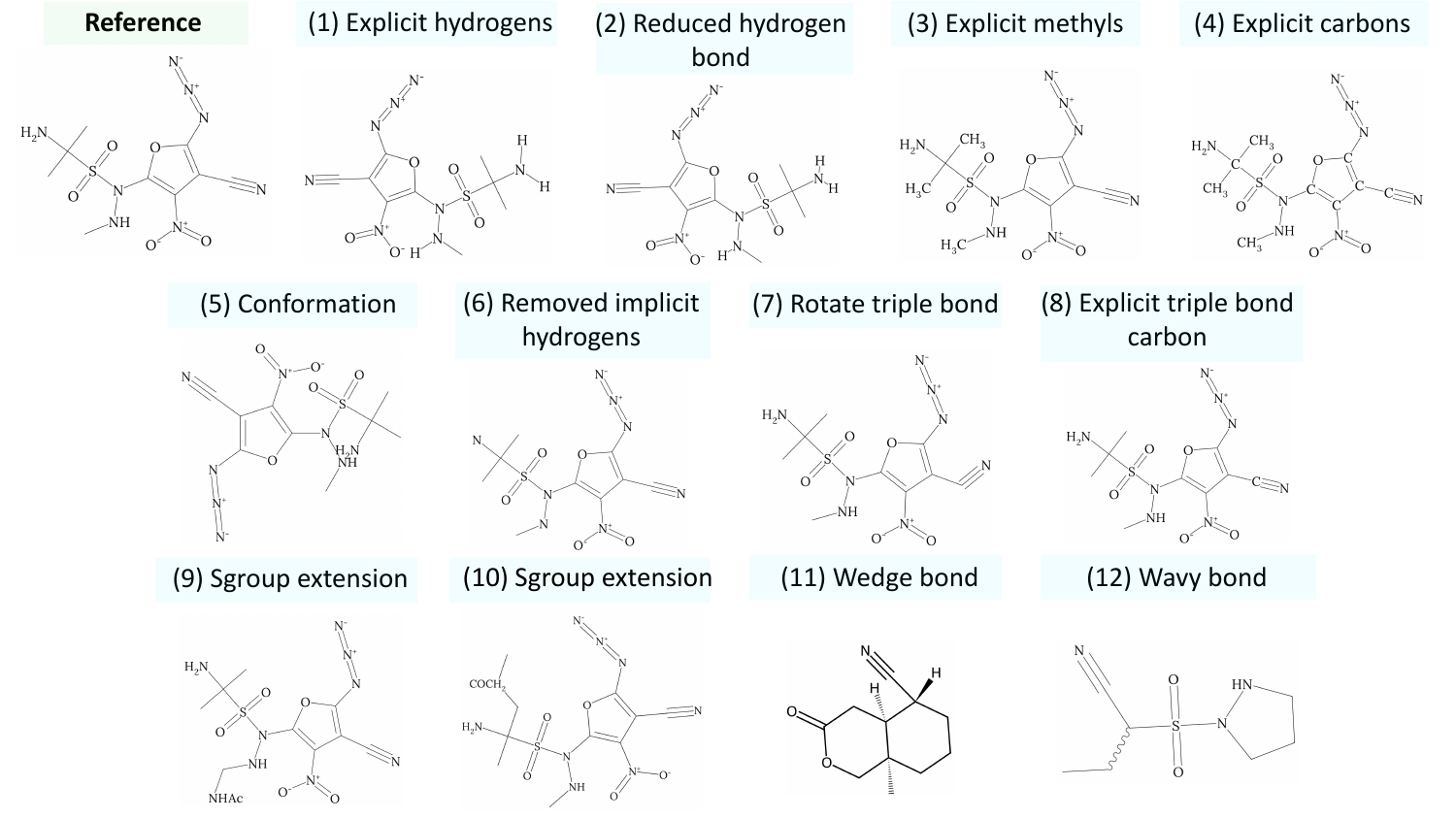}}
    \caption{\textbf{Molecule augmentations.} The figure illustrates all molecule augmentations applied independently on a reference molecule.}%
    \label{fig:molecule}\vspace{5mm}%
\end{figure*}

\begin{figure*}[]
\centering
\resizebox{\textwidth}{!}{
    \includegraphics*[trim={0 0cm 0 0cm}, clip, width=\textwidth]{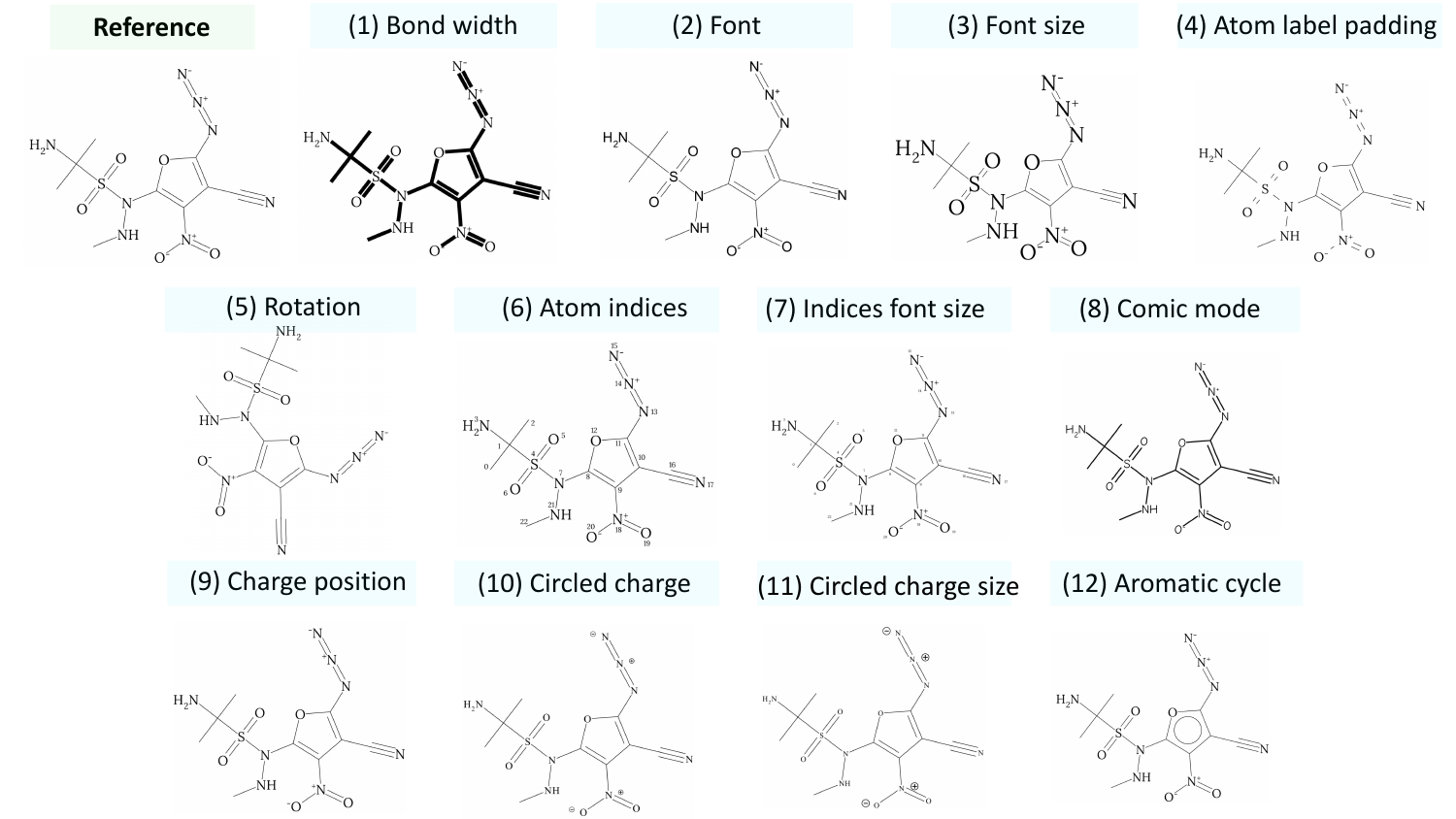}}
    \caption{\textbf{Rendering augmentations.} The figure describes all rendering augmentations applied independently on a reference drawing.}%
    \label{fig:rendering}\vspace{5mm}%
\end{figure*}

\begin{figure*}[]
\centering
    \resizebox{\textwidth}{!}{
    \includegraphics*[trim={0 0cm 0 0cm}, clip, width=\textwidth]{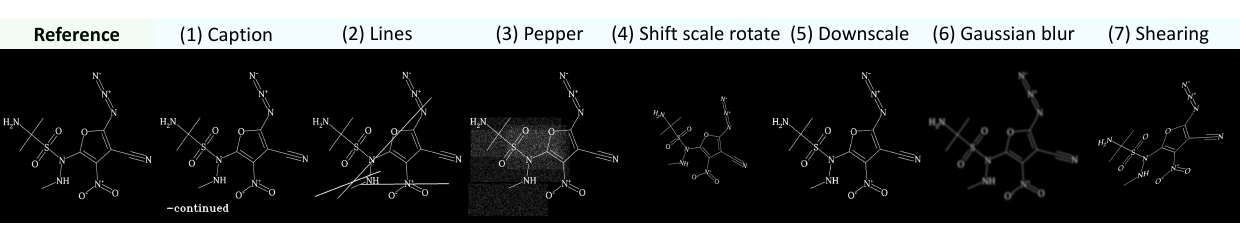}
    }
    \caption{\textbf{Image augmentations.} The figure illustrates all image augmentations applied independently on a reference image. Images are inverted in order to be compatible with the zero-padding used by default for convolutional layers.} 
    \label{fig:image}\vspace{0mm}%
\end{figure*}

In this section, we describe the synthetic training set augmentations as well as the generation of atom-level annotations. Contrary to the image captioning methods, our solution benefits from atom-level annotations, allowing to train with much less examples. In particular, MolGrapher is trained on 0.3 M images while Image2Graph \cite{Image2Graph} uses 7.1 M images, and DECIMER 2.0 \cite{DECIMER-AI} uses 450 M images.

\subsection{Augmentation}
The synthetic training set is augmented at the molecule, rendering and image levels.

\noindent\textbf{Molecule level.} As presented in \autoref{fig:molecule}, molecules are randomly transformed by: (1) displaying explicit hydrogens, (2) reducing of the size of bonds connected to explicit hydrogens, (3) displaying explicit methyls, (4) displaying explicit carbons, (5) selecting a molecular conformation, (6) removing implicit hydrogens of atom labels, (7) rotating triple bonds,  (8) displaying explicit carbons connected to triple bonds, adding artificial superatom groups with (9) single or (10) multiple attachment points, (11) displaying wedge bonds using solid or dashed bonds, and (12) displaying single bonds as wavy bonds. 

\noindent\textbf{Rendering level.} As demonstrated in \autoref{fig:rendering}, the rendering parameters used in RDKit \cite{RDKit} are randomly set: (1) the bond width, (2) the font, (3) the font size, (4) the atom label padding, (5) the molecule rotation, which does not rotate atom labels, (6) the display of atom indices and (7) their font size, (8) the hand-drawing style, (9) the charges positions, (10) the display of encircled charges and (11) their size, and (12) the display of aromatic cycles using circles.

\noindent\textbf{Image level.} As showcased in \autoref{fig:image}, images undergo several image augmentations on the fly: (1) the addition of random captions, (2) the addition of random lines, (3) the addition of pepper patches, (4) rotation, scaling, shifting, (5) resolution downscaling, (6) gaussian blurring and (7) x-y shearing. Finally, images are inverted in order to be compatible with the zero-padding used by default for convolutional layers.
The severity of augmentations is different for training the keypoint detector or the node classifier. Indeed, to only detect atoms positions, the keypoint detector does not need to precisely distinguish atom labels. 

\subsection{Atom-level Annotation}
Together with the training images, we generate the graph ground-truth, \ie the graph connectivity, the atoms and bonds labels, and their positions. RDKit allows to embed to the generated image some metadata, which stores the mapping between atom indices and their positions in the image. At the same time, we store a MolFile \cite{MolFile} containing the graph connectivity information and the class of each atom and bond. By combining both of them, we then create the graph ground-truth.

\begin{figure*}[t]
\centering
\resizebox{\textwidth}{!}{
    \includegraphics*[trim={0 4.4cm 0 5.2cm}, clip, width=\textwidth]{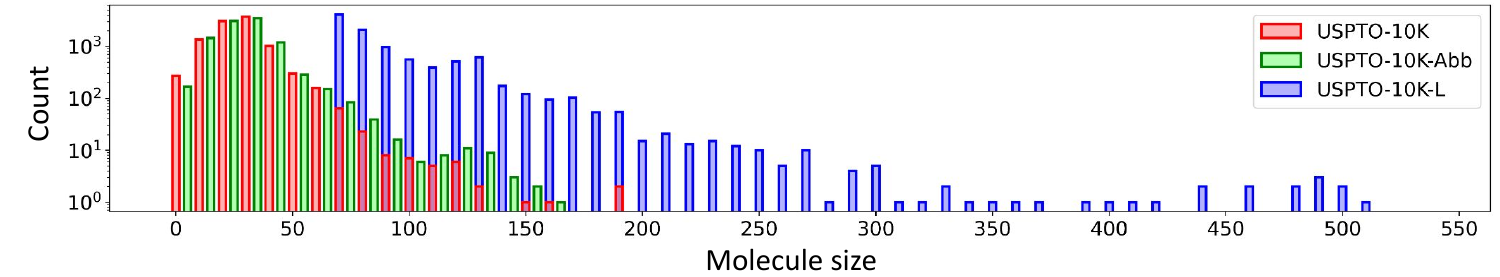}}
    \caption{\textbf{USPTO-30K molecule sizes distribution.} The figure illustrates the distribution of molecule sizes (number of atoms) in USPTO-10K (red), USPTO-10K-Abb (green) and USPTO-10K-L (blue). The population of each size bin is expressed in logarithmic scale.}
    \label{fig:size}\vspace{0mm}
\end{figure*}

\begin{figure*}[t] \vspace{0mm}
\centering
\resizebox{\textwidth}{!}{
    \includegraphics*[trim={0 3.6cm 0 5.2cm}, clip, width=\textwidth]{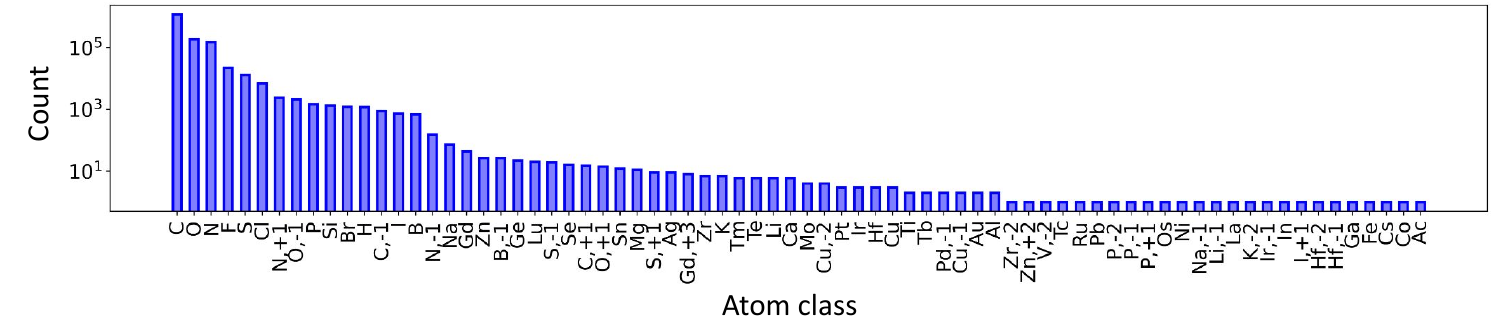}}
    \caption{\textbf{USPTO-30K atom classes distribution.} The figure shows the distribution of the atom classes in USPTO-30K. The count of each atom class is expressed in logarithmic scale.}
    \label{fig:atom}\vspace{0mm}
\end{figure*}

\begin{figure*}[] \vspace{0mm}
\centering
\resizebox{\textwidth}{!}{
    \includegraphics*[trim={0 3.6cm 0 5.2cm}, clip, width=\textwidth]{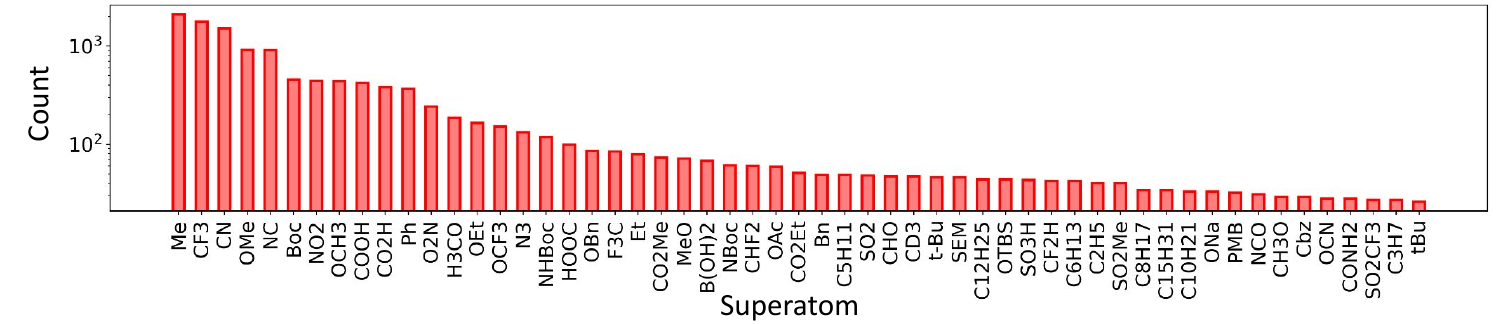}}
    \caption{\textbf{USPTO-10K-Abb abbreviations distribution.} The figure shows the distribution of the superatoms in USPTO-10K-Abb. Only superatoms with more than 25 occurrences are displayed. The count of each superatom is expressed in logarithmic scale.}
    \label{fig:sgroup}\vspace{0mm}
\end{figure*}

\begin{figure*}[!t]
\centering
    \resizebox{\textwidth}{!}{
    \includegraphics*[trim={0 1.5cm 0 1.5cm}, clip, width=\textwidth]{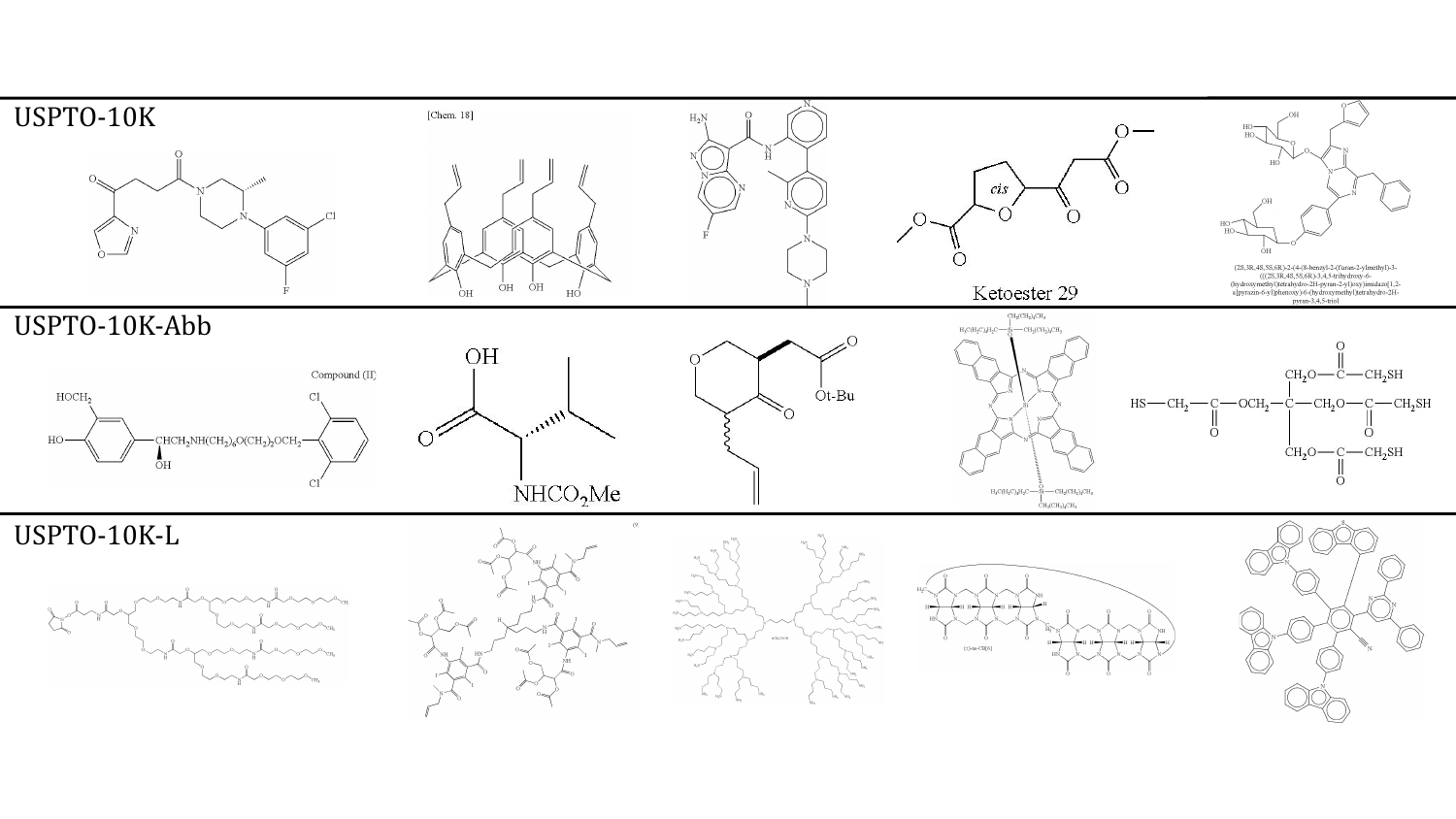}
    }
    \caption{\textbf{USPTO-30K example images.} The figure shows example images randomly selected in USPTO-10K, USPTO-10K-Abb and USPTO-10K-L.} 
    \label{fig:uspto30k}\vspace{0mm}%
\end{figure*}

\section{USPTO-30K Statistics and Examples}
\label{sec:uspto-30k}

In this section, we analyze the distribution of molecules in the dataset USPTO-30K, we compare its composition with other benchmarks, and visualize examples.

\subsection{Statistics}

\textbf{Molecule sizes.} \autoref{fig:size} presents the distribution of molecule sizes in USPTO-30K, the size of a molecule being defined as its number of atoms. 

\textbf{Atoms and superatom groups.} \autoref{fig:atom} and \autoref{fig:sgroup} show the distribution of atom classes and abbreviations in USPTO-30K. Randomly sampling images from all US patents from 2001 to 2020 allows us to cover a large diversity of atoms and superatom groups, including common but also exotic ones.

\textbf{Comparison with other benchmarks.} \autoref{tab:comparison} presents the comparison of the compositions of USPTO-30K and other benchmark datasets. USPTO-30K also contains more samples, disentangles the study of clean and abbreviated molecules. In comparison with existing sets, USPTO-30K contains more samples, a greater range of molecule sizes, more than $3\times$ as many atom classes, $2\times$ as many superatom classes, and a higher proportion of molecules with stereo-chemistry information. 

\subsection{Visualization}

\autoref{fig:uspto30k} illustrates some examples of images randomly sampled from USPTO-30K. 

\begin{figure*}[]
\centering
    \resizebox{\textwidth}{!}{
    \includegraphics*[trim={0 3cm 0 0cm}, clip, width=\textwidth]{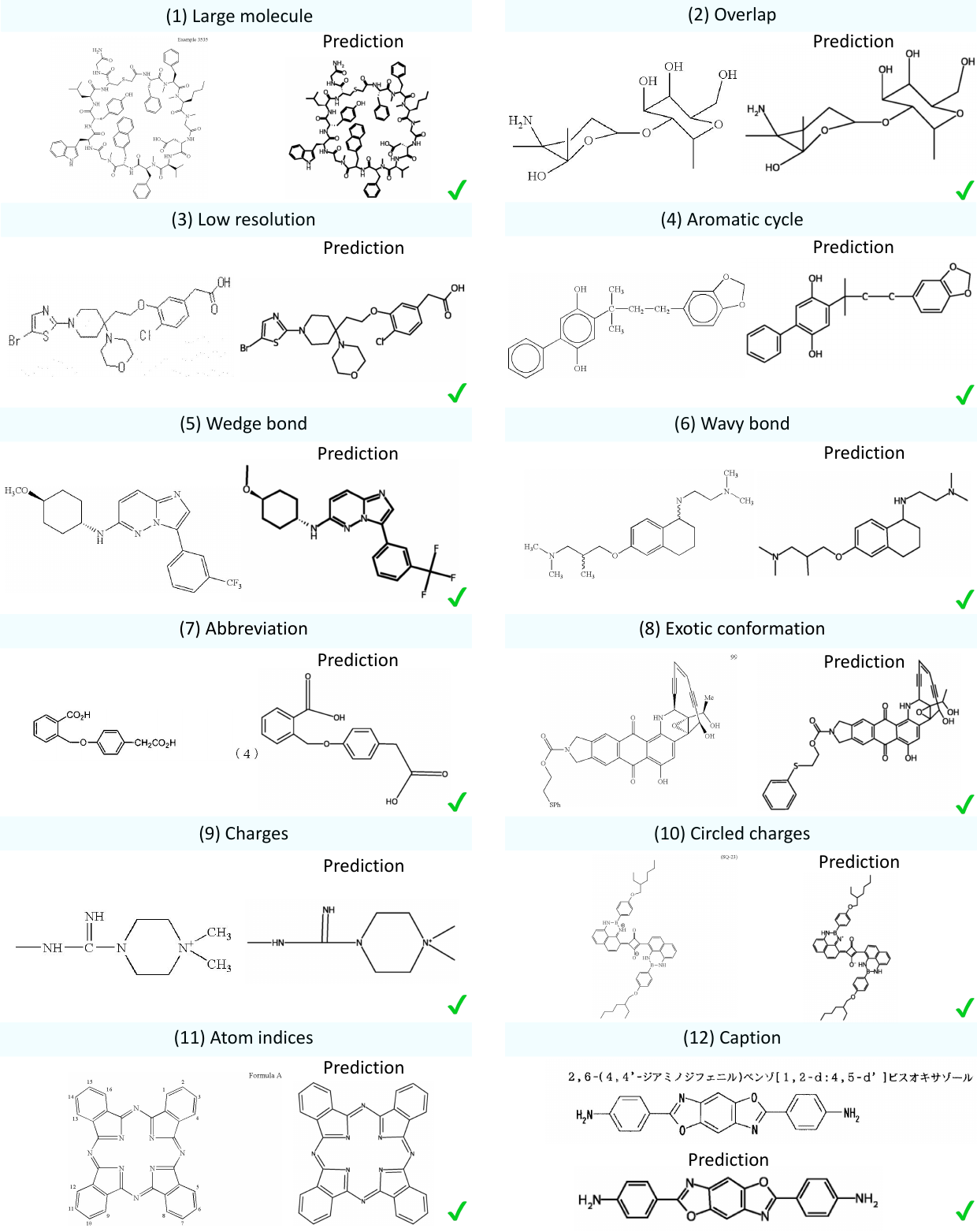}
    }
    \caption{\textbf{MolGrapher qualitative evaluation.} The figure shows MolGrapher predictions for a broad diversity of input molecule images. Input images (left) are displayed together with predicted molecules (right). The model can correctly handle challenging features, such as large molecule or overlapping bonds.}%
    \label{fig:show}\vspace{-4mm}%
\end{figure*}

\begin{figure*}[]
\centering
    \resizebox{\textwidth}{!}{
    \includegraphics*[trim={0 17.5cm 0 0cm}, clip, width=\textwidth]{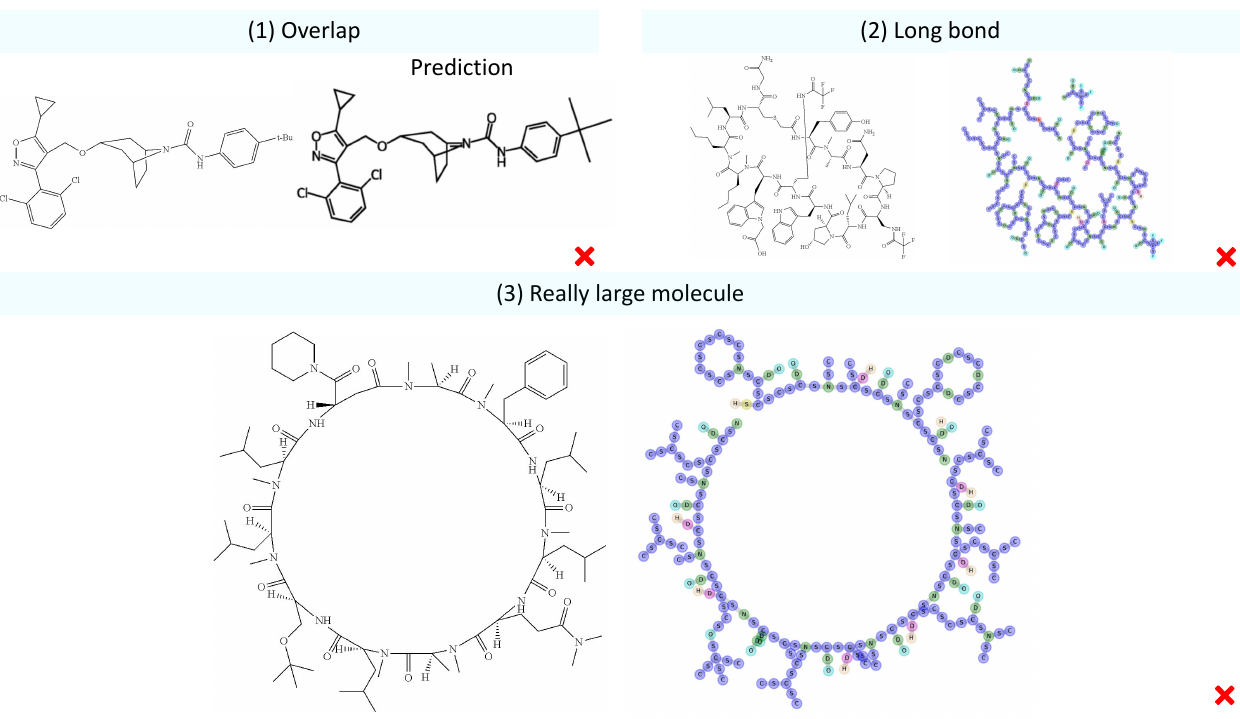}
    }
    \caption{\textbf{MolGrapher failure cases.} The figure illustrates some examples of failure cases of MolGrapher. Input images (left) are displayed together with predicted molecules (right). The predicted graph (right) is displayed when it can not be converted to a valid single-fragment molecule.}%
    \label{fig:failure}\vspace{1mm}%
\end{figure*}

\begin{figure*}[]
\centering
    \includegraphics*[trim={0 4.3cm 0 5.2cm}, clip, width=\textwidth]{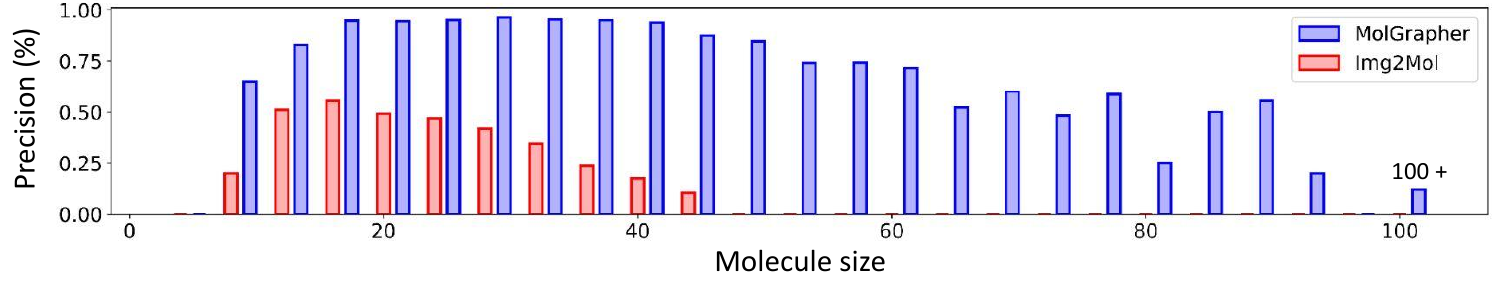}
    \caption{\textbf{USPTO-10K performance analysis.} The figure details the performance of MolGrapher (blue) and Img2Mol (red) on USPTO-10K with respect to the molecule size.}%
    \label{fig:size_performances}\vspace{-1mm}%
\end{figure*}

\section{Implementation details} 
\label{sec:implementation}

\textbf{Atom and bond classes.} 
The recognized atom classes are  `no atom', `C', `N', `O', `S', `F', `Cl', `P', `Br', `I', `B', `Si', `Sn', `Te', `Sb', `Bi', `Se', `Al', `As', `W', `Hg', `Ge', `In', `Na', `Pb', `Mg', `Pt', `Tl', `Fe', `Ru', `Cr', `Li', `Ar', `Pd', `Zr', `Zn', `Mo', `Xe', `U', `Po', `Ni', `K', `Cs', `At', `Yb', `Ti', `Tc' and `Os'. The recognized bond classes are `no bond', `single', `double', `triple',  `wedge-solid', `wedge-dashed' and `aromatic'.

\textbf{Superatom groups recognition.} Before applying PP-OCR \cite{PaddleOCR}, the molecule images are preprocessed by removing some of the bonds to simplify the text labels recognition. This is done by removing larger clusters of continuous filled pixels. Additionally, the mapping between recognized superatoms and submolecules is automatically created using the MolFiles provided by the United States Patent and Trademark Office (USPTO) \cite{USPTO}. In total, 819 abbreviations can be recognized, the most common ones being: `Me', `CF3', `CN', `NC', `OMe', `Boc', `OCH3', `NO2', `COOH', `Ph', `CO2H', `O2N', `H3CO', `OEt', `OCF3', `NHBoc', `N3', `Et', `HOOC', `OBn', `B(OH)2', `CHF2', `CO2Me', `F3C', `OAc' and `t-Bu'. Abbreviations also include R-groups labels such as  X', `Y', `Z', `R1', `R2` or `R10'. 

\textbf{Stereo-chemistry recognition.} In 2-D molecule depictions, stereo-chemistry is represented using solid or dashed wedge bonds, which respectively point towards or away from the viewer. To recognize wedge bonds, the model first predicts a node with a class `wedge-solid' or `wedge-dashed'. It is then needed to identify the direction of this bond. For this purpose, the ratio of filled pixels on both sides of the bond is computed. The side with the smallest amount of filled pixels is inside the depiction plane and the other one is outside. 

\textbf{Caption removal.} Although MolGrapher is trained with synthetic images containing random captions, it may not cover the complexity encountered during inference, such as Japanese captions. Thus, the OCR toolkit PP-OCR \cite{PaddleOCR} is used to detect and remove captions in images. Detected text cells which respect some criteria are replaced by a white background. For instance, if a recognized text sequence contains more than five consecutive lowercase characters, it should necessarily be a caption.  

\textbf{Keypoint refinement.} PP-OCR is also used to refine keypoint predictions for superatom groups having multiple attachment points, \ie connected to the rest of the molecule by multiple bonds. The keypoint detector learns to detect a keypoint at the extremities of each bond. For long abbreviated groups with multiple attachment points, each of its outgoing bond can be located at different positions along the abbreviation. This situation results in several detected keypoints for the same superatom. Therefore, if the initial prediction is an invalid molecule, PP-OCR is used to merge keypoints that are located within the same detected text cell.
\break
\textbf{Inference speed.} 
On a machine equipped with one NVIDIA A100 GPU and AMD EPYC 7763 CPU, processing images by batches of 20, MolGrapher annotates an average of 3.1 images per second. This runtime measurement was obtained by running MolGrapher on USPTO-30K.

\section{MolGrapher detailed analysis}
\label{sec:analysis}

\subsection{Qualitative evaluation}

\textbf{Qualitative examples.}
\autoref{fig:show} shows a sample of predictions for challenging input images. In practice, MolGrapher can be used to annotate low-quality images, unconventional drawings, images noised by additional information such as captions, atom indices, or stereo-chemistry annotations. MolGrapher recognizes abbreviated groups, stereo-chemistry and aromaticity. 

\textbf{Failure cases.}
A sample of failure cases is presented in \autoref{fig:failure}. In particular, example (1) is a molecule with challenging overlaps, leading to an incorrect double bond prediction. Example (2) shows a large molecule that contains a particularly long bond, which also overlaps with the molecule. This long bond is not detected because the supergraph construction algorithm discards it, according to its threshold of maximum bond length. Modifying the supergraph construction and generating bond overlaps, and long bond augmentations in the synthetic training, could mitigate these issues. Example (3) is an exceedingly large molecule with more than 110 atoms. Although most of the structure is correctly recognized, few bonds are occasionally missing. Further post-processing rules could be implemented to force the attachment of unconnected fragments.

\subsection{Quantitative evaluation}

\begin{table}[t]
\caption{\textbf{SOTA comparison with same training sets.} 
We evaluate the performance of MolGrapher by training it on various synthetic datasets of different sizes and comparing it to different methods. \textdagger: without hyper-parameters tuning. 
}%
\label{tab:training}%
\centering%
\resizebox{\linewidth}{!}{
\begin{tabular}{lccccc}
\toprule
\textbf{Method} & Synthetic training set & USPTO & Maybridge-UoB & \multicolumn{1}{c}{CLEF-2012} & \multicolumn{1}{c}{JPO} \\ \hline
CEDe & \begin{tabular}[c]{@{}c@{}}CEDe @ 10K \end{tabular} & 79.0 & 74.1 & 68.0 & 49.4 \\
\textbf{MolGrapher} & \begin{tabular}[c]{@{}c@{}}CEDe @ 10K \end{tabular} & \textbf{84.5} & \textbf{89.8} & \textbf{79.6} & \textbf{59.2} \\
\midrule
\textbf{MolGrapher} & \begin{tabular}[c]{@{}c@{}}MolGrapher @ 10K \end{tabular} & \textbf{87.5} & \textbf{93.8} & \textbf{86.2} & \textbf{61.5} \\
\midrule
Graph Generation \textsuperscript{\textdagger} & MolGrapher @ 300K & 31.3 & 71.6 & 25.3 & 24.2\\
Img2Mol \textsuperscript{\textdagger} & MolGrapher @ 300K & 15.5 & 23.2 & 10.0 & 11.3 \\
\textbf{MolGrapher} & MolGrapher @ 300K & \textbf{91.5} & \textbf{94.9} & \textbf{90.5} & \textbf{67.5} \\
\bottomrule
\end{tabular}}
\end{table}

\textbf{SOTA comparison on same training data.} To provide a comprehensive fair comparison with the available methods in \autoref{tab:training}. We trained MolGrapher on the synthetic training set of CEDe \cite{CEDe}. MolGrapher still outperforms CEDe on all benchmarks by a large margin. To assess the contribution of our training data, we also train MolGrapher on the same number (10k) of images as CEDe generated using our pipeline (resulting in $30\times$ less data). We then trained open-source methods on our full dataset (300k). Our method outperforms them by a substantial margin.

\textbf{Ablation study.}
We perform an ablation study on MolGrapher's components in \autoref{tab:modules}. 
In order to isolate the errors induced by each component, we replace the other two components with Ground-Truth (GT) oracle predictions.
We use a synthetic test set of 10k images, since the ground-truth graph with keypoint locations are not available on real datasets. 
We observe that the main source of errors is the keypoint detection, followed by the node classification.

\begin{table}[t]
\caption{\textbf{Ablation study of MolGrapher modules.} To isolate the errors stemming from each component, we replace the other two components with Ground-Truth (GT) oracle predictions.}
\label{tab:modules}
\centering
\resizebox{\linewidth}{!}{
\begin{tabular}{cccc}
\toprule
Keypoint detection & Node classification & Superatom recognition &  Synthetic test set @ 10K \\ \midrule
\textbf{Ours} & \textbf{Ours} & \textbf{Ours} & 94.3 \\
\textbf{Ours} & GT & GT & \textbf{95.8} \\
GT & \textbf{Ours} & GT & 97.2 \\
GT & GT & \textbf{Ours} & 98.1 \\
\bottomrule
\end{tabular}}
\end{table}

\subsection{Molecule Size Analysis}

\autoref{fig:size_performances} shows the precision of MolGrapher and Img2Mol \cite{Img2Mol} on USPTO-10K with respect to the molecule size. MolGrapher maintains a reasonable performance even for remarkably large molecules of 90 atoms. The locality of our approach allows to scale with respect to the molecule size. On the other hand, Img2Mol fails to recognize correctly any molecule of more than 50 atoms. 

\subsection{Error Prediction}

\begin{table}[t]
\centering% 
\caption{\textbf{MolGrapher error detection.} We evaluate the `detected error rate', \ie the proportion of MolGrapher errors that can be detected because the predicted graph is not convertible to a valid single-fragment molecule. The `filtered precision' is the precision computed on the filtered benchmarks, in which detected errors are not considered.}
\label{tab:error}
\resizebox{\linewidth}{!}{%
\begin{tabular}{lcccc}
\toprule
\multicolumn{1}{l}{MolGrapher} &\multicolumn{1}{l}{USPTO-10K} & USPTO-10K-Abb & \multicolumn{1}{l}{USPTO-10K-L} & \multicolumn{1}{l}{JPO}  \\ \hline
Detected error rate & 20.3 & 15.1 & 52.5 & 40.1 \\ 
\midrule
Precision & 93.3 & 82.8 & 31.3 & 67.5 \\
Filtered precision & \textbf{94.7} & \textbf{85.4} & \textbf{67.4} & \textbf{80.5} \\ \bottomrule
\end{tabular}}
\end{table}

Contrary to image captioning methods, which are trained to always generate valid SMILES sequences, our model identifies incorrect predictions in many cases. \autoref{tab:error} presents the `detected error rate' of MolGrapher, \ie the proportion of errors that can be detected among the total number of errors. Indeed, by performing low-level predictions, we can detect that the predicted graph is not convertible into a valid single-fragment molecule. \autoref{tab:error} also reports the `filtered precision', which is the precision of MolGrapher on the filtered benchmarks, in which the detected errors are not considered. To annotate scientific literature at large scale, and extract a viable source of knowledge, identifying incorrect predictions is critical.   

\section{Limitations and Future Works}

Currently, MolGrapher is unable to recognize markush structures, \ie depictions of sets of molecules using positional and frequency variation indications. As future work, we aim to generalize the MolGrapher graph structure to represent markush structures or even reactions.

{\small
\bibliographystyle{ieee_fullname}
\bibliography{egbib}
}

\end{document}